\documentclass[letterpaper, 10 pt, conference]{ieeeconf}  

\IEEEoverridecommandlockouts                              

\usepackage{amsmath}
\usepackage{graphicx}
\usepackage{amssymb} 

\usepackage{booktabs}   
\usepackage{multirow}   
\usepackage{makecell}   
\usepackage{graphicx}   

\usepackage{xcolor}

\newcommand{\best}[1]{\textcolor{red}{\textbf{#1}}}
\newcommand{\second}[1]{\textcolor{blue}{\underline{#1}}}
\newcommand{\NA}{--}

\title{\LARGE \bf
From LiDAR Maps to Visual Localization: Unified Visual Association for Robust Point-Line-Plane Pose Estimation
}

\author{Wentao Zhao, Zikun Chen, Yihe Niu, Haoyu Chen, Jingchuan Wang$^{*},~\textit{Senior Member, IEEE}$ 
\thanks{
Wentao Zhao, Zikun Chen, and Jingchuan Wang are with the School of Automation and Intelligent Sensing, Institute of Medical Robotics, Shanghai Jiao Tong University.
Yihe Niu is with the School of Mathematical Sciences Shanghai Jiao Tong University.
Haoyu Chen is with the School of Electronic and Information Engineering, Beijing Jiaotong University.
}
}

\begin{document}
\bstctlcite{IEEEexample:BSTcontrol}

\maketitle
\thispagestyle{empty}
\pagestyle{empty}

\begin{abstract}
Camera localization in a prior LiDAR map provides a persistent geometric
reference for long-term robotic navigation, yet remains challenging because
of the substantial modality gap between camera images and point-cloud maps.
We present a unified localization framework that makes the LiDAR map
\emph{visually addressable} rather than relying on a dedicated image--LiDAR
correspondence model.
Map geometry and reflectivity are rendered into LiDAR-derived quasi-images
with explicit 2D--3D provenance, enabling camera observations and rendered
map views to share mature visual features and matchers for both global
localization and continuous pose tracking.
Point and line correspondences are established through this common visual
interface, while the retained provenance recovers metric LiDAR geometry and
line-supported planar constraints for pose estimation.
To improve robustness under ambiguous associations and weak geometry, we
further introduce a distribution-aware, observability-complementary
optimization strategy.
Instead of reducing matching ambiguity to a scalar confidence, candidate
association distributions are propagated into directional pose-information
uncertainty, and reliable structural factors are selectively reinforced
according to their ability to complement the currently weak pose directions.
Experiments on the EuRoC MAV benchmark and self-collected real-world sequences
demonstrate accurate global localization and robust continuous 6-DoF tracking
using only a pre-built LiDAR map as the persistent prior, including under
severe illumination variations and dynamic occlusions.
\end{abstract}

\section{INTRODUCTION}

Accurate camera localization in a prior 3D map is essential for autonomous
robots operating over long trajectories.
LiDAR maps are attractive persistent priors because they provide metrically
accurate and illumination-invariant geometry.
Camera localization against such maps involves two related but distinct
problems: \emph{global localization}, which determines the map-relative pose
without a valid initialization, and \emph{continuous pose tracking}, which
maintains this pose once localization has been established.
Existing approaches generally bridge the image--LiDAR gap in two ways.
One line relies on camera-derived appearance, such as reference RGB images,
visual descriptors, or colorized/textured maps, or learns dedicated
cross-modal correspondence models whose transferability may depend on the
training domain.
Another line directly exploits LiDAR geometry through registration or
structural primitives~\cite{wolcott2014visual,ye2020dsl}, but cannot directly
benefit from the mature local descriptors and matchers available in the image
domain, particularly for discriminative point-level association.
These limitations make it difficult to support both global localization and
continuous tracking directly from a LiDAR prior without camera-derived map
appearance.

In this work, we take a different approach: rather than learning a dedicated
image--LiDAR correspondence model, we make the LiDAR map
\emph{visually addressable}.
Given a viewpoint, map geometry and reflectivity are rendered into a
LiDAR-derived quasi-image with explicit provenance to the underlying 3D
points.
Camera observations and rendered map views can therefore share mature visual
feature extractors and matchers without modality-specific correspondence
training, while the retained 2D--3D provenance preserves metric LiDAR geometry
for pose estimation.
For global localization, multi-heading virtual map views are rendered offline
to construct the retrieval database directly from the LiDAR prior, without an
additional visual mapping traversal.
For continuous tracking, the predicted camera pose instead renders a local
pose-conditioned quasi-image.
Point and line correspondences are established through the shared visual
interface, while line-supported LiDAR geometry further provides planar
constraints.
The resulting point, line, and plane factors retain their native geometric
residuals during optimization.

Bridging the modality gap alone, however, does not guarantee reliable pose
estimation.
Point, line, and plane measurements become weak along different motion
directions and therefore provide complementary geometric constraints
~\cite{yang2019observability,zheng2025tcviml}.
At the same time, repetitive structures and weak appearance can produce
multiple plausible map associations.
Importantly, association ambiguity and geometric weakness are not equivalent:
two visually ambiguous candidates may induce nearly identical pose
information, whereas a confident correspondence may contribute little along a
currently weak pose direction.
Existing methods typically address correspondence reliability and geometric
observability separately~\cite{zheng2025safety}.
This leads to a more relevant question for map-based estimation:
\emph{how much reliable pose information does an uncertain correspondence
provide along the directions that are currently weak?}

We address this question by propagating association distributions into
directional pose information.
A local observability analysis first reveals the complementary weak directions
of point, line, and plane constraints.
For practical estimation, weak 6-DoF pose modes are then identified from the
current information matrix.
Rather than collapsing a set of plausible associations into a scalar matching
confidence, we evaluate the pose-information contribution induced by each
candidate and model its variation across the association distribution.
This yields a conservative estimate of reliable directional information,
allowing line and plane factors to be selectively reinforced only when they
provide consistent information that complements the current weak pose modes.
Together with temporal visual and inertial constraints, this results in a
unified framework for global localization and continuous 6-DoF pose tracking
using a LiDAR map as the persistent prior.

Our main contributions are:
\textbf{1)} We introduce a visualizable LiDAR-map representation that makes a
geometry-and-reflectivity prior directly accessible to mature visual features
and matchers.
LiDAR-derived quasi-images provide a common interface for global localization
and continuous tracking without modality-specific correspondence training,
while enabling the global database to be constructed without reference imagery
or an additional visual mapping traversal.
\textbf{2)} We develop a heterogeneous point--line--plane estimation scheme in
which point and line associations share the same visual front-end, while
explicit 2D--3D provenance recovers metric LiDAR geometry and line-supported
planar constraints for their native geometric residuals.
\textbf{3)} We propose a distribution-aware, observability-complementary
optimization strategy that propagates association ambiguity into directional
pose-information uncertainty and selectively reinforces reliable structural
factors according to their contribution to weak pose directions.
\textbf{4)} Extensive benchmark and real-world robotic experiments validate
accurate global localization and robust tracking under challenging motion,
illumination changes, and dynamic occlusions.

\section{RELATED WORK}

\begin{table}[t]
\centering
\caption{Representative prior-map localization approaches.
G/T denote global localization/tracking.}
\label{tab:related_work}
\setlength{\tabcolsep}{2.8pt}
\renewcommand{\arraystretch}{1.00}

\resizebox{\linewidth}{!}{%
\begin{tabular}{l|c|l|l}
\toprule
Method & Task & Strategy & Requirement \\
\midrule

\makecell[l]{HLoc, Surfel Reloc.
\cite{sarlin2019coarse,ye20213d}}
& \makecell[tc]{G}
& Visual-assisted matching
& Camera-derived reference \\

\makecell[l]{LIP-Loc, Poses as Queries
\cite{liploc,posesasqueries}}
& \makecell[tc]{G}
& Cross-modal learning
& Modality-specific training \\

\makecell[l]{I2D-Loc, I2D-Loc++
\cite{chen2022i2d,yu2024i2dpp}}
& \makecell[tc]{T}
& RGB--depth matching
& Modality-specific training \\

\makecell[tl]{2D--3D, PPL, TC-VIML,\\
Plane Loc.~\cite{yu2020monocular,zheng2025safety,
zheng2025tcviml,han2025plane}}
& \makecell[tc]{T}
& \makecell[tl]{Line/plane structural\\matching}
& \makecell[tl]{Reliable structural\\support} \\

\midrule

\makecell[tl]{Ours}
& \makecell[tc]{G+T}
& \makecell[tl]{Quasi-image + visual\\PLP association}
& \makecell[tl]{No reference imagery or\\cross-modal training} \\

\bottomrule
\end{tabular}%
}
\vspace{-5mm}

\end{table}

\noindent\textbf{Global Localization in Prior 3D Maps.}
Global localization estimates the camera pose in a prior map without a valid
initialization.
Conventional visual relocalization methods retrieve candidate views and recover
metric poses through geometric verification.
ORB-SLAM~\cite{mur2015orb}, HLoc~\cite{sarlin2019coarse}, and point--line
visual maps~\cite{xu2025airslam} achieve accurate localization but require
camera-derived appearance or visual descriptors in the prior representation.
For LiDAR-based priors, surfel relocalization~\cite{ye20213d} still relies on
reference image sequences, while LIP-Loc~\cite{liploc} and
Poses as Queries~\cite{posesasqueries} learn dedicated image--LiDAR
representations for cross-modal retrieval or pose estimation.

\noindent\textbf{Continuous Pose Tracking in Prior LiDAR Maps.}
Without a prior map, visual and visual--inertial systems
~\cite{leutenegger2015keyframe,qin2018vins,teed2021droid,
xu2025airslam,otplvio} provide accurate motion estimation and may reduce
accumulated drift through loop closure, but lack continuous anchoring to a
pre-built persistent map.
With a valid map-relative initialization, prior-map methods exploit
reflectivity alignment~\cite{wolcott2014visual}, point-cloud or surfel
representations~\cite{zuo2019visual,ye2020dsl}, Gaussian mixtures
~\cite{zuo2020multimodal}, or learned correspondences from rendered depth and
intensity maps~\cite{chen2022i2d,yu2024i2dpp,shen2024intensity}.
Structural approaches further employ lines~\cite{yu2020monocular}, planes
~\cite{han2025plane}, reliability modeling~\cite{zheng2025safety}, and
observability-aware prior-line fusion~\cite{zheng2025tcviml}.

Overall, existing methods either rely on camera-derived map appearance,
learn dedicated cross-modal representations, or match geometric structures
shared by both modalities, as summarized in Table~\ref{tab:related_work}.
The first two require additional visual map information or modality-specific
training, while geometry-centric approaches depend strongly on the quality
and availability of structural primitives.
Our method instead renders the LiDAR prior into quasi-images, enabling mature
visual point--line matching while retaining metric 2D--3D provenance for PLP
constraints.

\section{METHOD}

\subsection{System Overview and Problem Formulation}

\begin{figure}[t]
    \centering
    \includegraphics[width=\linewidth]{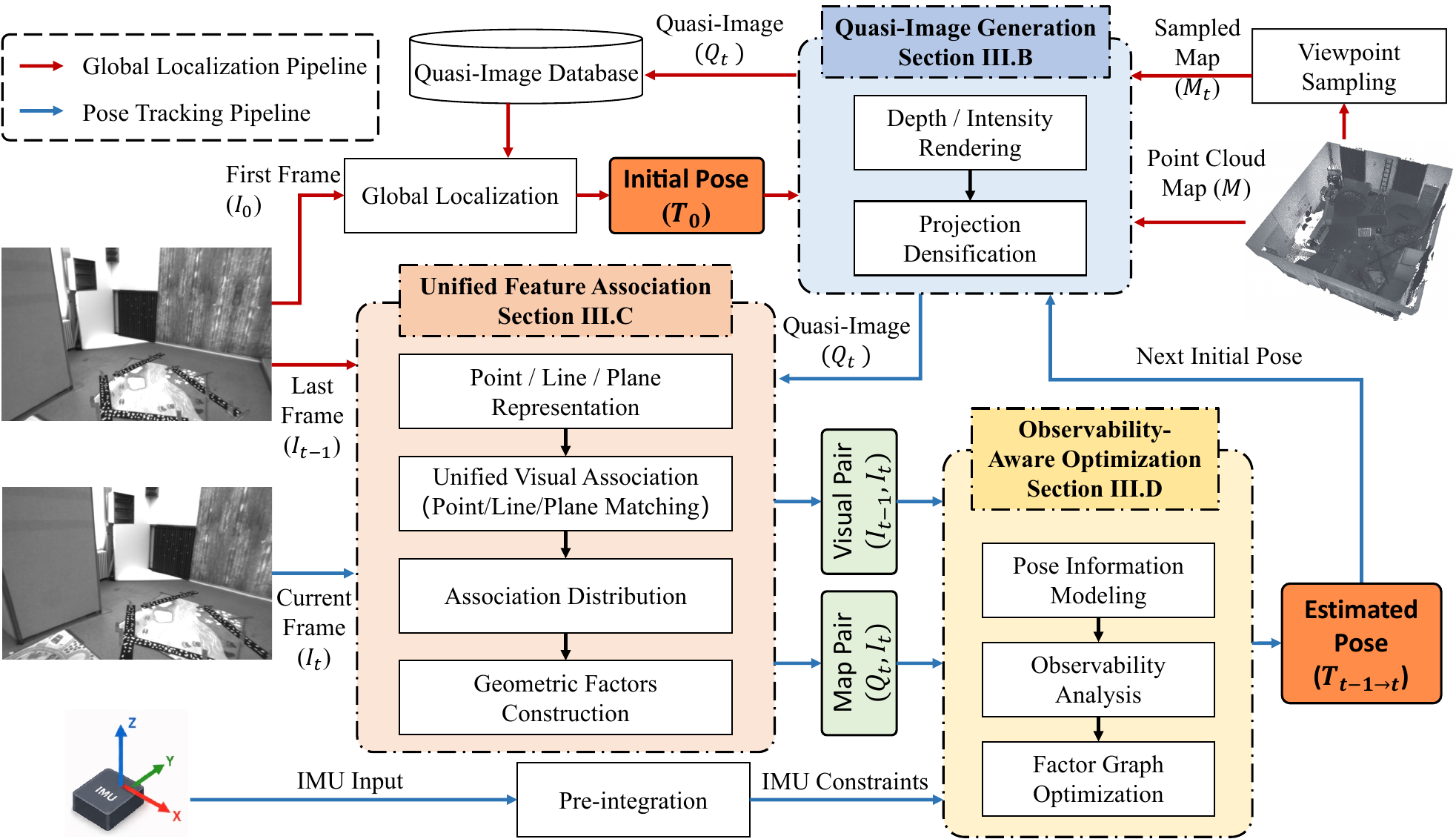}
    \caption{Overview of the proposed localization framework. Global
    localization initializes the camera pose from an offline quasi-image
    database, while continuous tracking jointly exploits visual--visual,
    visual--map, and inertial constraints.}
    \label{fig:overview}
    \vspace{-5mm}
\end{figure}

\noindent\textbf{Problem formulation.}
We consider localization in a pre-built LiDAR map
\begin{equation}
    \mathcal{M}
    =
    \left\{
    \left(\mathbf{P}_j^M,\rho_j\right)
    \right\}_{j=1}^{N_M},
\end{equation}
where $\mathbf{P}_j^M$ and $\rho_j$ denote the 3D position and LiDAR
reflectivity of the $j$-th map point.
Let $\mathcal{I}_t=(I_t^L,I_t^R)$ denote the stereo observation at time $t$.
We consider
\begin{equation}
    (\mathcal{I}_0,\mathcal{M})
    \rightarrow \mathbf{T}_0,
    \qquad
    (\mathcal{I}_{0:t},\mathcal{U}_{0:t},\mathcal{M},\mathbf{T}_0)
    \rightarrow
    \mathcal{T}_{0:t},
    \label{eq:localization_tasks}
\end{equation}
where the two mappings denote global localization and continuous
stereo--inertial tracking, respectively, and
$\mathcal{T}_{0:t}=\{\mathbf{T}_k\}_{k=0}^{t}$ with
$\mathbf{P}^{C_t}=\mathbf{T}_t\mathbf{P}^{M}$.

\noindent\textbf{Global localization and continuous tracking.}
As shown in Fig.~\ref{fig:overview}, representative map viewpoints are sampled
offline to construct a quasi-image database.
For global localization, visual retrieval and geometric verification over the
offline quasi-image database estimate the initial pose $\mathbf{T}_0$.
During tracking, the predicted pose $\hat{\mathbf{T}}_t$ renders a local
pose-conditioned quasi-image $Q_t$.
The shared visual front-end establishes temporal and map associations from
\begin{equation}
    \mathcal{P}_t^{V}=(I_{t-1},I_t),
    \qquad
    \mathcal{P}_t^{M}=(Q_t,I_t),
    \label{eq:visual_map_pairs}
\end{equation}
which are jointly optimized with stereo and IMU constraints to estimate the
current map-relative pose.

\subsection{Quasi-Image Generation}

The prior LiDAR map is rendered into a visual-compatible quasi-image for both
offline database construction and online pose-conditioned tracking.

\noindent\textbf{Depth--Intensity Rendering.}
Given a rendering pose $\mathbf{T}$, each map point
$(\mathbf{P}_i^M,\rho_i)$ is projected onto the virtual camera as
\begin{equation}
    \mathbf{u}_i=\pi(\mathbf{T}\mathbf{P}_i^M), \qquad
    D_s(\mathbf{u}_i)=z_i,\quad
    I_s(\mathbf{u}_i)=\tilde{\rho}_i ,
\end{equation}
where $D_s$ and $I_s$ denote the sparse depth and normalized reflectivity
maps, respectively.
Visibility conflicts are resolved by retaining the nearest projected point.

\noindent\textbf{Depth--Intensity Cross-Guided Completion.}
Direct projection produces sparse and locally discontinuous observations.
We therefore perform cross-guided neighborhood aggregation, where depth
continuity constrains reflectivity propagation and reflectivity consistency
guides depth completion.
For the valid neighborhoods $\mathcal{N}_I(\mathbf{x})$ and
$\mathcal{N}_D(\mathbf{x})$ selected by the complementary modality,
\begin{equation}
    \hat I(\mathbf{x})
    =\max_{\mathbf{y}\in\mathcal{N}_I(\mathbf{x})} I_s(\mathbf{y}),
    \quad
    \hat D(\mathbf{x})
    =\frac{1}{|\mathcal{N}_D(\mathbf{x})|}
    \sum_{\mathbf{y}\in\mathcal{N}_D(\mathbf{x})}D_s(\mathbf{y}).
\end{equation}
Max aggregation preserves salient reflectivity responses, whereas average
aggregation improves local depth continuity without propagating across
inconsistent structures.

Following the pseudo-color representation in~\cite{wang2025salt}, the
completed reflectivity and depth cues are jointly encoded as
\begin{equation}
\begin{aligned}
    H &= h(\hat I), \qquad S=s_0,\\
    V &= \beta_1+\beta_2 h\!\left(\mathcal{F}(\hat D)\right),
    \qquad
    Q=\mathcal{C}(H,S,V),
\end{aligned}
\label{eq:quasi_encoding}
\end{equation}
where $h(\cdot)$ denotes intensity normalization, $\mathcal{F}(\cdot)$
extracts local depth variations, and $\mathcal{C}(\cdot)$ converts the
pseudo-color representation into the quasi-image.
Reflectivity provides appearance-like contrast, while depth variations
emphasize geometric boundaries.
During online tracking, the predicted pose $\hat{\mathbf{T}}_t$ determines
the rendering viewpoint and yields the pose-conditioned quasi-image $Q_t$.

\begin{figure}[t]
    \centering
    \includegraphics[width=0.8\linewidth]{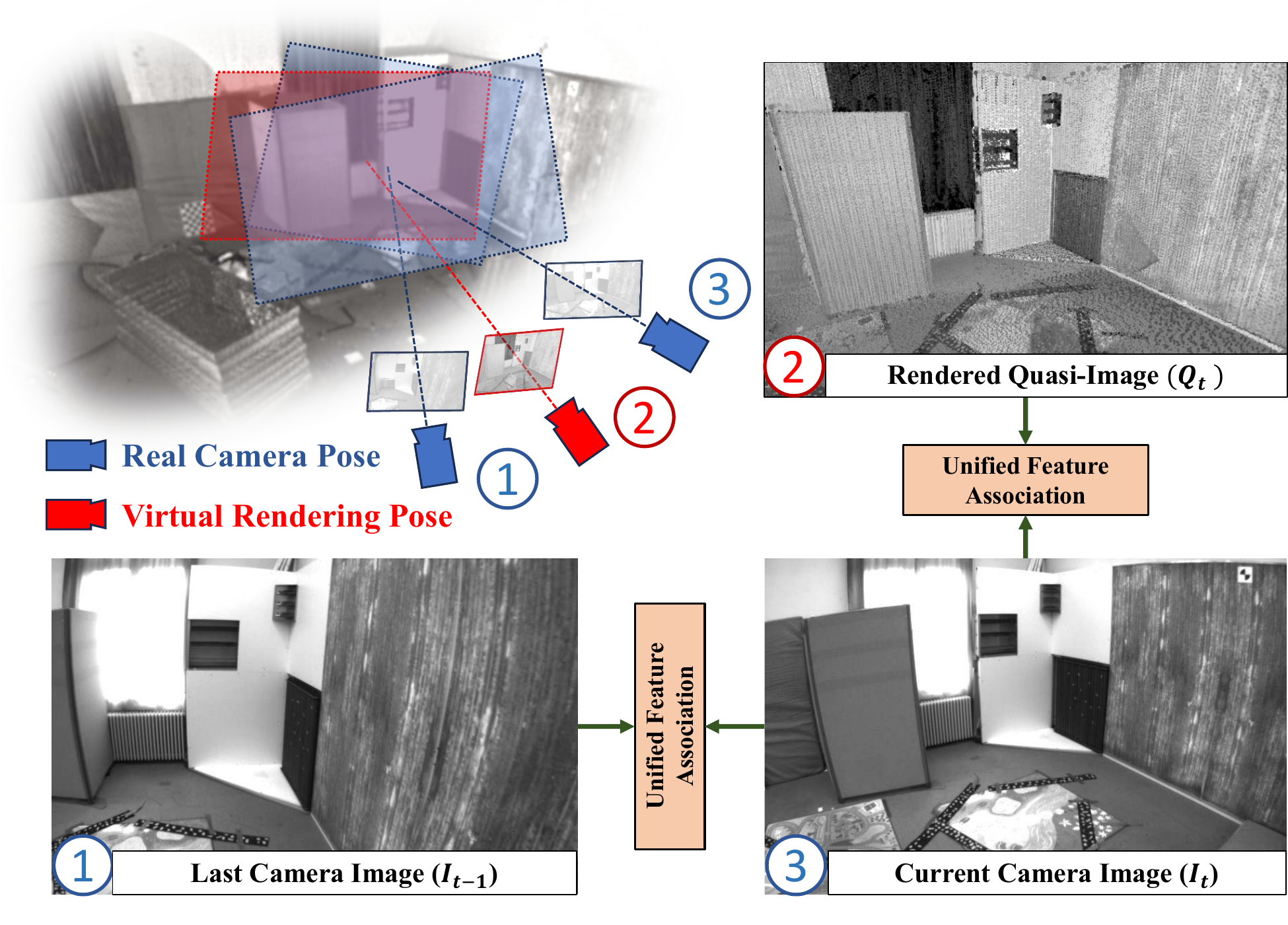}
    \caption{
    Unified visual association between real camera views ($I_{t-1}, I_t$) and
    the LiDAR-rendered quasi-image $Q_t$; blue and red frustums denote real and
    virtual rendering poses, respectively.
    }
    \label{fig:quasi_image}
    \vspace{-5mm}
\end{figure}

\subsection{Unified Feature Association}

As shown in Fig.~\ref{fig:quasi_image}, the previous and current camera
images are acquired from real camera viewpoints, whereas $Q_t$ is rendered
from a virtual map viewpoint.
The visual pair $\mathcal{P}_t^{V}$ and the visual--map pair
$\mathcal{P}_t^{M}$ are processed by the same source-agnostic visual
front-end to establish point, line, and plane (PLP) constraints.

\noindent\textbf{Unified PLP Representation.}
We adopt the point--line front-end of OTPL-VIO~\cite{otplvio} and share a
frozen pretrained PL-Net between camera images and quasi-images, without
modality-specific fine-tuning.
For each input, PL-Net detects keypoints and line segments, and provides
per-keypoint descriptors together with point- and line-specific feature maps
$\mathbf{F}_{p}$ and $\mathbf{F}_{l}$.
Point features directly retain their learned descriptors, whereas an explicit
descriptor is constructed for each detected line segment by sampling both
feature maps along the segment.
For a line $l_i$ with uniformly sampled locations
$\{\mathbf{s}_{ik}\}_{k=1}^{N_s}$, we compute
\begin{equation}
\begin{aligned}
    \bar{\mathbf{f}}_{i}^{b}
    &=
    \mathcal{N}\!\left(
    \frac{1}{N_s}\sum_{k=1}^{N_s}
    \mathbf{F}_{b}(\mathbf{s}_{ik})
    \right),
    \quad b\in\{p,l\},\\
    \mathbf{f}_{i}^{l}
    &=
    \mathcal{N}\!\left(
    [\,\gamma_l\bar{\mathbf{f}}_{i}^{l};
    \gamma_p\bar{\mathbf{f}}_{i}^{p}\,]
    \right).
\end{aligned}
\label{eq:hybrid_line_descriptor}
\end{equation}
where $\mathcal{N}(\cdot)$ denotes $\ell_2$ normalization and
$\gamma_l,\gamma_p$ balance line-structural and point-context information.
Thus, the line representation aggregates features from both point and line
branches along the complete segment rather than relying on discrete point
correspondences.

Planar constraints are constructed from geometrically consistent
line-supported structures.
For quasi-images, the corresponding pixels retain explicit 2D--3D
provenance and are traced back to the original LiDAR points, from which
metric planar geometry is recovered.

\noindent\textbf{Source-Agnostic Visual Association.}
The same association pipeline is applied to both
$(I_{t-1},I_t)$ and $(Q_t,I_t)$.
Point correspondences are obtained from the PL-Net point descriptors using
LightGlue, while line segments are matched independently using their hybrid
segment descriptors through entropy-regularized optimal transport
~\cite{otplvio}.
For line sets $\mathcal{L}^{A}$ and $\mathcal{L}^{B}$, the descriptor cost
between segments $i$ and $j$ is
\begin{equation}
    C_{ij}^{l}
    =
    1-
    \left\langle
    \mathbf{f}_{i}^{l,A},
    \mathbf{f}_{j}^{l,B}
    \right\rangle .
    \label{eq:line_matching_cost}
\end{equation}
For visual--map associations, multiple plausible map hypotheses are retained
before hard selection.
For feature $i$ of type $g\in\{p,l,\pi\}$, we represent them by
\begin{equation}
    \mathcal{A}_i^g
    =
    \left\{
    (c,p_{ic}^g)
    \mid c\in\mathcal{C}_i^g
    \right\},
    \qquad
    \sum_{c\in\mathcal{C}_i^g}p_{ic}^g=1 ,
    \label{eq:association_distribution}
\end{equation}
where $c$ indexes a candidate map hypothesis and $p_{ic}^g$ is its normalized
association score.
Point and line hypotheses are obtained from alternative quasi-image
associations and lifted through the retained 2D--3D provenance to metric
LiDAR geometry, while planar hypotheses are induced from geometrically
consistent line-supported LiDAR structures.
For each structural hypothesis, its supporting LiDAR samples jointly define
one candidate rather than independent associations.
The highest-probability hypothesis provides the nominal correspondence, while
the full distribution is retained for subsequent pose-information modeling.

\noindent\textbf{Distribution-Aware Factor Construction.}
The retained 2D--3D provenance lifts visual--map correspondences back to
metric LiDAR geometry, while the three primitives preserve their native
geometric residuals:
\begin{equation}
\begin{gathered}
    \mathbf{r}_i^{p}
    =
    \mathbf{u}_i-\pi\!\left(\mathbf{T}_t\mathbf{P}_i^M\right),
    \qquad
    r_{im}^{l}
    =
    {\mathbf{l}_i}^{\top}
    \pi\!\left(\mathbf{T}_t\mathbf{P}_{im}^{M}\right),\\
    r_{im}^{\pi}
    =
    {\mathbf{n}_i^M}^{\top}
    \mathbf{T}_t^{-1}\mathbf{P}_{im}^{C_t}
    +d_i^M .
\end{gathered}
\label{eq:plp_residuals}
\end{equation}
They correspond to point reprojection, point-to-line, and point-to-plane
constraints, respectively.
Here, $\mathbf{P}_{im}^{M}$ denotes LiDAR support associated with the matched
map line, while the map plane $(\mathbf{n}_i^M,d_i^M)$ is fitted from
non-degenerate LiDAR support induced by the corresponding line-based planar
structure.
The query-side support $\mathbf{P}_{im}^{C_t}$ is obtained from the local
visual--inertial reconstruction.
Rather than collapsing $\mathcal{A}_i^g$ into a scalar matching confidence,
we retain the candidate distribution and evaluate the pose information
induced by alternative geometric hypotheses in the following optimization.

\subsection{Observability-Aware Optimization}

\noindent\textbf{Pose Information Modeling.}
Point, line, and plane measurements do not become weak along the same motion
directions.
Therefore, our goal is not simply to introduce more structural constraints,
but to determine whether a candidate factor provides reliable information
along the pose directions that are currently weak.
We first analyze the local geometric complementarity of the three primitives
and then extend this insight to the full 6-DoF pose-information space.

For factor $i$ and association hypothesis $c\in\mathcal{C}_i$, let
$\mathbf{J}_{ic}$ denote the Jacobian of its native residual and
$\mathbf{W}_{i}^{\mathrm{meas}}$ its measurement weight.
The corresponding candidate-specific pose information is
\begin{equation}
    \mathbf{H}_{ic}
    =
    \mathbf{J}_{ic}^{\top}
    \mathbf{W}_{i}^{\mathrm{meas}}
    \mathbf{J}_{ic}.
    \label{eq:candidate_information}
\end{equation}
Measurement uncertainty is modeled by
$\mathbf{W}_{i}^{\mathrm{meas}}$, whereas association uncertainty is retained
separately through the candidate distribution in
(\ref{eq:association_distribution}).
For brevity, the primitive superscript is omitted in the following.

To understand the geometric sensitivity of different primitives, let
$\boldsymbol{\ell}$ denote the viewing ray of a point,
$\mathbf{m}_{L}$ the normal of the interpretation plane induced by an image
line, and $\mathbf{n}$ the normal of a fitted 3D plane.
Their local Jacobians can be written as
\begin{equation}
\begin{gathered}
    \mathbf{J}_{P}
    =
    \mathbf{J}_{\pi}(\mathbf{P})
    \bigl[[\mathbf{P}]_{\times}\!-\!\mathbf{I}\bigr],
    \quad
    \mathbf{J}_{L,i}
    =
    \tfrac{1}{Z_i}
    \bigl[(\mathbf{m}_{L}\!\times\!\mathbf{P}_i)^{\top}
    \!-\!\mathbf{m}_{L}^{\top}\bigr],
    \\
    \mathbf{J}_{\Pi,i}
    =
    \bigl[(\mathbf{n}\!\times\!\mathbf{Q}_i)^{\top}
    \!-\!\mathbf{n}^{\top}\bigr].
\end{gathered}
\label{eq:plp_jacobians}
\end{equation}
The local sensitivities of the three primitive observations exhibit the
following translational and rotational weak directions:
\begin{equation}
\begin{alignedat}{3}
    \mathcal{N}_{t}^{P} &= \operatorname{span}(\boldsymbol{\ell}), \;&
    \mathcal{N}_{t}^{L} &= \mathbf{m}_{L}^{\perp}, \;&
    \mathcal{N}_{t}^{\Pi} &= \mathbf{n}^{\perp},\\
    \mathcal{N}_{R}^{P} &= \operatorname{span}(\boldsymbol{\ell}), \;&
    \mathcal{N}_{R}^{L} &= \operatorname{span}(\mathbf{m}_{L}), \;&
    \mathcal{N}_{R}^{\Pi} &= \operatorname{span}(\mathbf{n}).
\end{alignedat}
\label{eq:plp_weak_directions}
\end{equation}
These directions characterize local primitive sensitivity rather than the
null space of the complete estimator. Point and line translational information
decays as $O(1/Z^2)$, whereas plane-normal translation is not attenuated by
perspective depth.

\noindent\textbf{Observability-Complementary Reweighting.}
We next show why the heterogeneous geometric primitives can complement
different weak pose directions.
Consider a non-degenerate local structure $P\in L\subset\Pi$, where the scene
plane does not pass through the camera center and the 3D line has a
non-degenerate image projection.
Let $\mathbf{v}_{L}$ denote the 3D line direction.
The geometry satisfies
\begin{equation}
    \mathbf{m}_{L}^{\top}\boldsymbol{\ell}=0,
    \qquad
    \mathbf{v}_{L}\parallel
    \mathbf{m}_{L}\times\mathbf{n}.
    \label{eq:plp_geometric_relation}
\end{equation}
The first relation implies that point and line factors retain the same
translational weak direction along the viewing ray.
The second shows that the common translational weak direction of line and
plane factors is the 3D line direction.
More explicitly,
\begin{equation}
\begin{aligned}
    \mathcal{N}_{t}^{P}\cap\mathcal{N}_{t}^{L}
    &=\operatorname{span}(\boldsymbol{\ell}),
    &
    \boldsymbol{\ell}^{\top}
    \mathbf{H}_{tt,\Pi}\boldsymbol{\ell}
    &>0,\\
    \mathcal{N}_{t}^{L}\cap\mathcal{N}_{t}^{\Pi}
    &=\operatorname{span}(\mathbf{v}_{L}),
    &
    \mathbf{v}_{L}^{\top}
    \mathbf{H}_{tt,P}\mathbf{v}_{L}
    &>0 ,
\end{aligned}
\label{eq:plp_complementarity}
\end{equation}
where $\mathbf{H}_{tt}$ denotes the translational information block.
The first inequality follows from
$\mathbf{n}^{\top}\boldsymbol{\ell}\neq0$ for a visible point on a plane not
passing through the camera center.
The second follows from
$\mathbf{v}_{L}\not\parallel\boldsymbol{\ell}$ for a non-degenerate projected
line.

Hence, under non-degenerate local geometry,
\begin{equation}
    \mathcal{N}_{t}^{P}
    \cap
    \mathcal{N}_{t}^{L}
    \cap
    \mathcal{N}_{t}^{\Pi}
    =
    \{\mathbf{0}\}.
    \label{eq:plp_joint_translation}
\end{equation}
Thus, the three primitives have no common translational null direction:
the residual point--line weakness can be complemented by the plane, whereas
the residual line--plane weakness can be complemented by the point.
This translational analysis provides the geometric intuition for combining
heterogeneous PLP constraints.
It does not imply that the complete estimator is exactly degenerate; in
practice, the weak modes generally lie in the full 6-DoF pose space and are
identified directly from the current pose-information matrix.

We therefore identify the actual weak modes from the current base information.
Let $\mathbf{H}_{\mathrm{imu}}$, $\mathbf{H}_{V}$, and $\mathbf{H}_{P}$
denote the $6\times6$ current-pose information contributions induced by the
inertial, temporal visual, and map-point factors after linearization.
The base pose information is
\begin{equation}
    \mathbf{H}_{0}
    =
    \mathbf{H}_{\mathrm{imu}}
    +
    \mathbf{H}_{V}
    +
    \mathbf{H}_{P}.
    \label{eq:base_information}
\end{equation}
Since rotational and translational perturbations have different physical
scales, directly comparing the eigenvalues of $\mathbf{H}_{0}$ may introduce
an artificial dependence on the pose parameterization.
We therefore perform the observability analysis in a normalized pose space.
For the local perturbation
$\delta\boldsymbol{\xi}
=
[\delta\boldsymbol{\theta}^{\top},
 \delta\mathbf{t}^{\top}]^{\top}$,
let $L_t$ denote a characteristic scene length, chosen as the median depth of
the valid map correspondences, and define
\begin{equation}
    \mathbf{D}_{t}
    =
    \operatorname{diag}
    \left(
        \mathbf{I}_{3},
        L_t\mathbf{I}_{3}
    \right).
    \label{eq:pose_normalization}
\end{equation}
The normalized base information is then
\begin{equation}
    \widetilde{\mathbf{H}}_{0}
    =
    \mathbf{D}_{t}^{\top}
    \mathbf{H}_{0}
    \mathbf{D}_{t}
    =
    \mathbf{U}\boldsymbol{\Lambda}\mathbf{U}^{\top},
    \label{eq:normalized_base_information}
\end{equation}
where
$\boldsymbol{\Lambda}
=
\operatorname{diag}(\lambda_1,\ldots,\lambda_6)$
and $\mathbf{u}_k$ denotes the $k$-th normalized pose-information direction.
This normalization is used only for observability analysis and structural
reweighting; the original geometric residuals and state parameterization are
retained in the subsequent optimization.

We define a continuous weakness coefficient
\begin{equation}
    a_k
    =
    \left[
        1-\frac{\sqrt{\lambda_k}}{\tau}
    \right]_{+},
    \label{eq:weakness}
\end{equation}
where $[\cdot]_{+}=\max(0,\cdot)$.
Thus, directions with smaller information eigenvalues receive larger weakness
coefficients, whereas sufficiently constrained directions are not emphasized.

The next question is how much a plausible structural association can improve
a particular weak mode.
For consistency with the above analysis, the candidate-specific information
in~\eqref{eq:candidate_information} is expressed in the same normalized pose
space as
\begin{equation}
    \widetilde{\mathbf{H}}_{ic}
    =
    \mathbf{D}_{t}^{\top}
    \mathbf{H}_{ic}
    \mathbf{D}_{t}.
    \label{eq:normalized_candidate_information}
\end{equation}
Consider the perturbed normalized information
\begin{equation}
    \widetilde{\mathbf{H}}(\alpha)
    =
    \widetilde{\mathbf{H}}_{0}
    +
    \alpha\widetilde{\mathbf{H}}_{ic}.
\end{equation}
For a simple eigenvalue, first-order eigenvalue perturbation gives
\begin{equation}
    g_{ick}
    \triangleq
    \left.
    \frac{
        d\lambda_k
        \bigl(\widetilde{\mathbf{H}}(\alpha)\bigr)
    }{d\alpha}
    \right|_{\alpha=0}
    =
    \mathbf{u}_{k}^{\top}
    \widetilde{\mathbf{H}}_{ic}
    \mathbf{u}_{k}.
    \label{eq:candidate_directional_information}
\end{equation}
Therefore, $g_{ick}$ is not an ad-hoc confidence score, but the first-order
increase of the $k$-th normalized pose-information eigenvalue contributed by
candidate $c$.
It directly measures whether the candidate provides information along the
directions that are currently weak.

Association ambiguity introduces a second issue.
A visually ambiguous correspondence is not necessarily geometrically
ambiguous: different plausible candidates may induce nearly identical pose
information.
We therefore propagate the complete association distribution into the
directional-information statistics
\begin{equation}
\begin{aligned}
    \mu_{ik}
    &=
    \sum_{c\in\mathcal{C}_i}p_{ic}g_{ick},
    \qquad
    \sigma_{ik}^{2}
    =
    \sum_{c\in\mathcal{C}_i}
    p_{ic}(g_{ick}-\mu_{ik})^{2},\\
    \hat g_{ik}
    &=
    [\mu_{ik}-\kappa\sigma_{ik}]_{+}.
\end{aligned}
\label{eq:directional_information_distribution}
\end{equation}
Here, $\mu_{ik}$ measures the expected contribution along direction
$\mathbf{u}_k$, while $\sigma_{ik}$ measures how sensitive this contribution
is to alternative associations.
Importantly, if all plausible candidates induce the same directional
information, i.e.,
$g_{ick}=g_{ik}$ for all $c\in\mathcal{C}_i$, then
$\sigma_{ik}=0$ regardless of how diffuse the association probabilities are.
Thus, association ambiguity does not necessarily imply directional
pose-information uncertainty.

Finally, let
\begin{equation}
    \widetilde{\bar{\mathbf{H}}}_{i}
    =
    \sum_{c\in\mathcal{C}_i}
    p_{ic}\widetilde{\mathbf{H}}_{ic}
    \label{eq:expected_information}
\end{equation}
denote the expected normalized information of factor $i$.
For structural factors $i\in\{L,\Pi\}$, we define
\begin{equation}
    q_i
    =
    \frac{
        \sum_{k=1}^{6}a_k\hat g_{ik}
    }{
        \operatorname{tr}
        (\widetilde{\bar{\mathbf{H}}}_{i})+\epsilon
    },
    \qquad
    \mathbf{W}_{i}^{*}
    =
    (1+\beta q_i)\mathbf{W}_{i}^{\mathrm{meas}}.
    \label{eq:complementary_reweighting}
\end{equation}
The numerator measures the reliable information supplied along the currently
weak pose directions, while the denominator normalizes this contribution by
the factor's overall pose-information magnitude.
A large $q_i$ therefore requires both reliable association-induced information
and strong complementarity to what the estimator currently lacks.
In this way, structural factors are selectively reinforced according to their
reliable contribution to weak pose modes, rather than according to matching
confidence or total information magnitude alone.

\noindent\textbf{Factor Graph Optimization.}
The current state is estimated by jointly minimizing inertial, temporal
visual, and visual--map factors:
\begin{equation}
    \mathbf{X}^{*}
    =
    \arg\min_{\mathbf{X}}
    \left(
        E_{\mathrm{imu}}
        +E_{V}
        +E_{P}
        +E_{L}^{*}
        +E_{\Pi}^{*}
    \right).
    \label{eq:final_optimization}
\end{equation}
At each outer iteration, the factors are first linearized at the current state
estimate.
The base information matrix is then formed to identify the weak pose modes,
after which candidate directional information and complementarity scores are
evaluated.
The resulting structural weights are fixed during the subsequent
Gauss--Newton update and recomputed after relinearization.

\section{EXPERIMENTS}

\subsection{Experimental Setup}

\begin{figure}[t]
    \centering
    \includegraphics[width=0.8\linewidth]{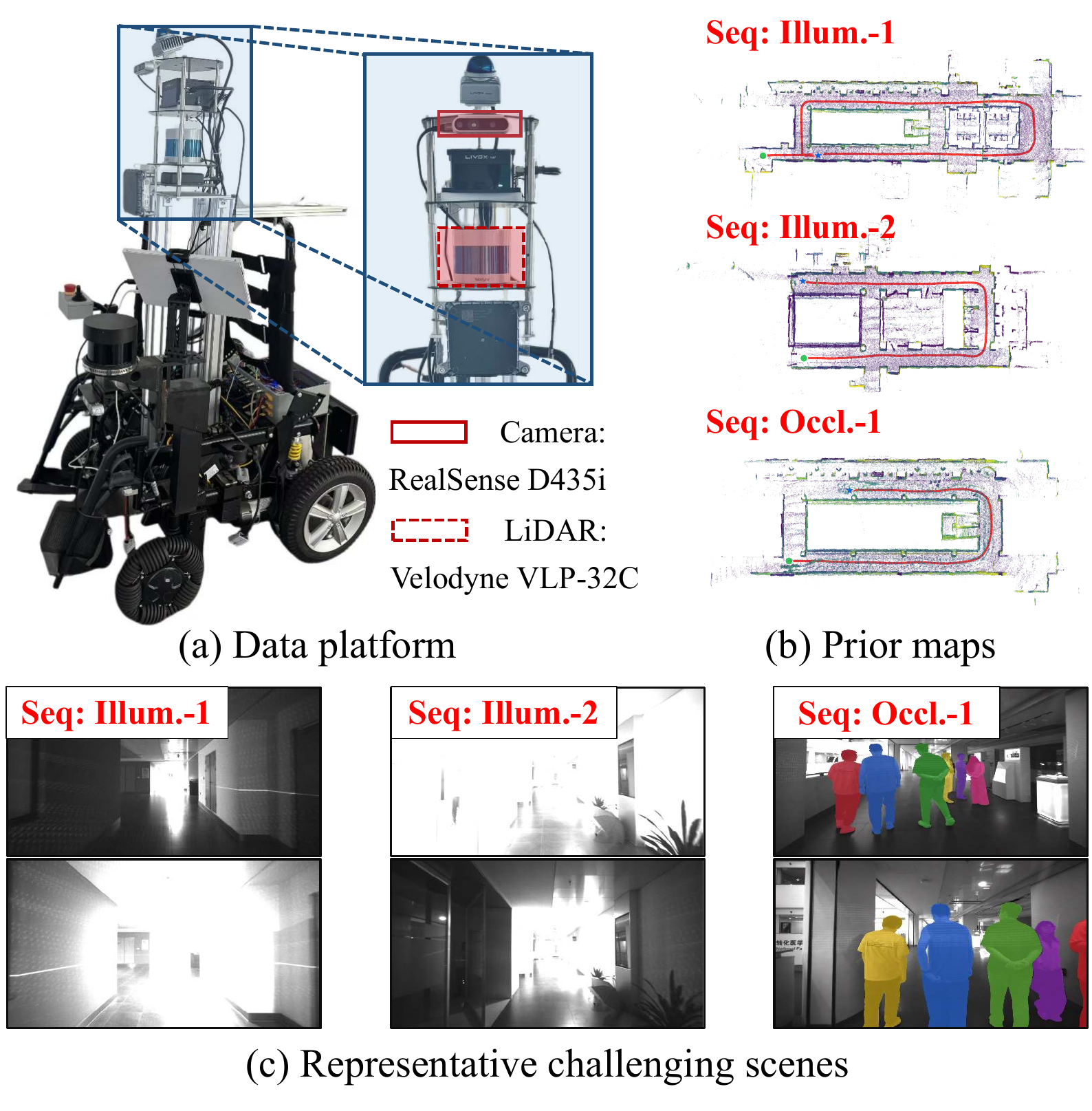}
    \vspace{-2mm}
    \caption{
    Real-world experimental setup and representative challenging scenes.
    (a) Mobile robotic platform equipped with an Intel RealSense D435i
    camera--IMU and a 32-beam Velodyne LiDAR.
    (b) Prior LiDAR maps of the three self-collected sequences:
    \emph{Illum.-1}, \emph{Illum.-2}, and \emph{Occl.-1}.
    (c) Representative camera observations showing severe illumination
    changes and sustained pedestrian occlusions.
    }
    \label{fig:real_platform}
    \vspace{-5mm}
\end{figure}

\noindent\textbf{Datasets and Platforms.}
We evaluate the proposed framework on the EuRoC MAV \cite{burri2016euroc} benchmark and
self-collected indoor sequences.
EuRoC V1 and V2 provide rapid motion and illumination
variations.
Our self-collected sequences further target structural degradation, illumination
changes, and dynamic occlusions.
Data are captured by a mobile robot with a RealSense D435i and a 32-beam
Velodyne LiDAR (Fig.~\ref{fig:real_platform}).
Prior maps and reference trajectories are generated offline by LiDAR--inertial
odometry with loop closure and dynamic-object filtering; online localization
uses only stereo images and IMU.
All three self-collected sequences contain weak or degraded structural support;
\emph{Illum.-1} and \emph{Illum.-2} additionally exhibit substantial
illumination changes, while \emph{Occl.-1} contains sustained pedestrian
occlusions over large portions of the static scene.

\noindent\textbf{Baselines and Metrics.}
For global localization on EuRoC, we compare with representative
single-image relocalization methods, including ORB-SLAM
~\cite{mur2015orb}, hierarchical localization (NV+SP)
~\cite{sarlin2019coarse}, surfel-based relocalization
~\cite{ye20213d}, and AirSLAM~\cite{xu2025airslam}.
The first three use published results under the same cross-sequence protocol,
while AirSLAM is evaluated using our implementation.
Global localization is evaluated by localization recall and the mean
translation error of the relocalized poses.

For continuous pose tracking, we compare with representative prior-free visual
and visual--inertial baselines
~\cite{qin2018vins,teed2021droid,leutenegger2015keyframe,
xu2025airslam,otplvio}, as well as prior-map localization methods covering
point-, line-, plane-, and dense-map representations
~\cite{zuo2019visual,ye2020dsl,zuo2020multimodal,yu2020monocular,
zheng2025safety,zheng2025tcviml,han2025plane}.
Our method directly uses a pre-built LiDAR map as the persistent prior,
without requiring camera-derived appearance to be attached to the map.
Pose tracking is evaluated using ATE RMSE.

\noindent\textbf{Implementation Details.}
For global localization, multi-heading virtual viewpoints are sampled offline
to construct a quasi-image database with visual descriptors and 2D--3D
provenance.
Given a query image, BoW retrieval recalls candidate views, which are re-ranked
using LightGlue correspondences and lifted through the retained provenance to
obtain 2D--3D matches.
Multi-view voting aggregates consistent correspondences across nearby views,
followed by multi-hypothesis PnP verification to estimate the global pose.
We evaluate two configurations with different retrieval budgets while keeping
the remaining pipeline unchanged.
For continuous pose tracking, the predicted pose renders a local quasi-image,
and the resulting PLP map constraints are jointly optimized with temporal
visual and IMU pre-integration factors.

\begin{table}[t]
\centering
\caption{Global relocalization on the EuRoC MAV dataset.
Each entry reports Recall [\%] / mATE [cm].}
\label{tab:euroc_global}
\setlength{\tabcolsep}{4.2pt}
\renewcommand{\arraystretch}{1.10}
\resizebox{\linewidth}{!}{%
\begin{tabular}{l|ccccc}
\toprule
Method
& V101 & V103 & V201 & V203 & Avg. \\
\midrule

ORB-SLAM~\cite{mur2015orb}
& 55.8 / \second{3.11}
& 17.6 / 13.28
& 61.5 / \second{2.96}
& 1.0 / 9.85
& 34.0 / 7.30 \\

HLoc (NV+SP)~\cite{sarlin2019coarse}
& \best{99.9} / \best{2.50}
& \best{93.8} / 20.44
& \best{98.9} / 5.54
& \best{88.7} / 32.39
& \best{95.3} / 15.22 \\

Surfel Reloc.~\cite{ye20213d}
& \second{93.5} / 4.91
& 71.5 / \best{5.25}
& 88.3 / \best{2.82}
& 53.5 / \best{5.20}
& 76.7 / \best{4.55} \\

AirSLAM~\cite{xu2025airslam}
& 93.1 / 7.60 
& 83.8 / 10.18 
& 86.6 / 10.45 
& 60.0 / 10.63
& 80.9 / 9.72 \\

\midrule
Ours (Fast)
& 79.3 / 15.34
& 87.6 / 8.09
& 82.1 / 8.80
& 72.2 / 11.32
& 80.3 / 10.89 \\

Ours (Full)
& 93.1 / 7.29 
& \second{90.5} / \second{7.96} 
& \second{90.2} / 8.22 
& \second{79.1} / \second{8.90}
& \second{88.2} / \second{8.09} \\
\bottomrule
\end{tabular}}
\vspace{-3mm}
\end{table}

\begin{table}[t]
\centering
\caption{ATE RMSE [cm] on the EuRoC MAV dataset.
Methods are grouped according to whether a prior map is used.}
\label{tab:euroc_tracking}
\setlength{\tabcolsep}{3.6pt}
\renewcommand{\arraystretch}{1.08}

\resizebox{\linewidth}{!}{%
\begin{tabular}{c|l|ccc|ccc}
\toprule
\multicolumn{2}{c|}{Method}
& V101 & V102 & V103
& V201 & V202 & V203 \\
\midrule

\multirow{5}{*}{\rotatebox{90}{w/o Map}}
& VINS-Fusion~\cite{qin2018vins}
& 10.2 & 9.9 & 11.2
& 11.0 & 12.4 & 25.2 \\

& DROID-SLAM~\cite{teed2021droid}
& 10.3 & 16.5 & 15.8
& 10.2 & 11.5 & 20.4 \\

& OKVIS~\cite{leutenegger2015keyframe}
& 4.0 & 6.7 & 12.0
& 5.5 & 15.0 & 24.0 \\

& AirSLAM~\cite{xu2025airslam}
& 3.3 & 3.2 & 23.8
& 3.6 & 8.3 & 16.8 \\

& OTPL-VIO~\cite{otplvio}
& 3.8 & 6.8 & 6.9
& 3.3 & 8.3 & 12.0 \\

\midrule

\multirow{9}{*}{\rotatebox{90}{w/ Map}}
& Zuo et al.~\cite{zuo2019visual}
& 5.6 & 5.5 & 8.7
& 6.9 & 8.9 & 14.9 \\

& DSL~\cite{ye2020dsl}
& 3.5 & 3.4 & 4.5
& 2.6 & \second{2.3} & \second{10.3} \\

& GMMLoc~\cite{zuo2020multimodal}
& \second{3.0} & \second{2.3} & \second{4.0}
& \second{1.7} & 3.6 & \NA \\

& 2D--3D~\cite{yu2020monocular}
& 8.9 & 6.9 & 17.3
& 16.6 & 13.2 & 63.5 \\

& PPL~\cite{zheng2025safety}
& 15.3 & 9.2 & 18.6
& 18.5 & 15.5 & 31.9 \\

& PPL-OR~\cite{zheng2025safety}
& 15.2 & 9.9 & 16.1
& 16.6 & 13.6 & 31.2 \\

& TC-VIML~\cite{zheng2025tcviml}
& 6.8 & 8.4 & 18.2
& 6.8 & 10.8 & 25.4 \\

& Plane-based Loc.~\cite{han2025plane}
& \best{1.6} & \second{2.3} & \second{4.0}
& \second{1.7} & 3.6 & \NA \\

\cmidrule{2-8}

& Ours
& 3.3
& \best{2.1}
& \best{2.6}
& \best{1.4}
& \best{1.5}
& \best{3.2} \\

\bottomrule
\end{tabular}%
}
\vspace{-5mm}
\end{table}

\begin{figure}[t]
    \centering
    \includegraphics[width=\linewidth]{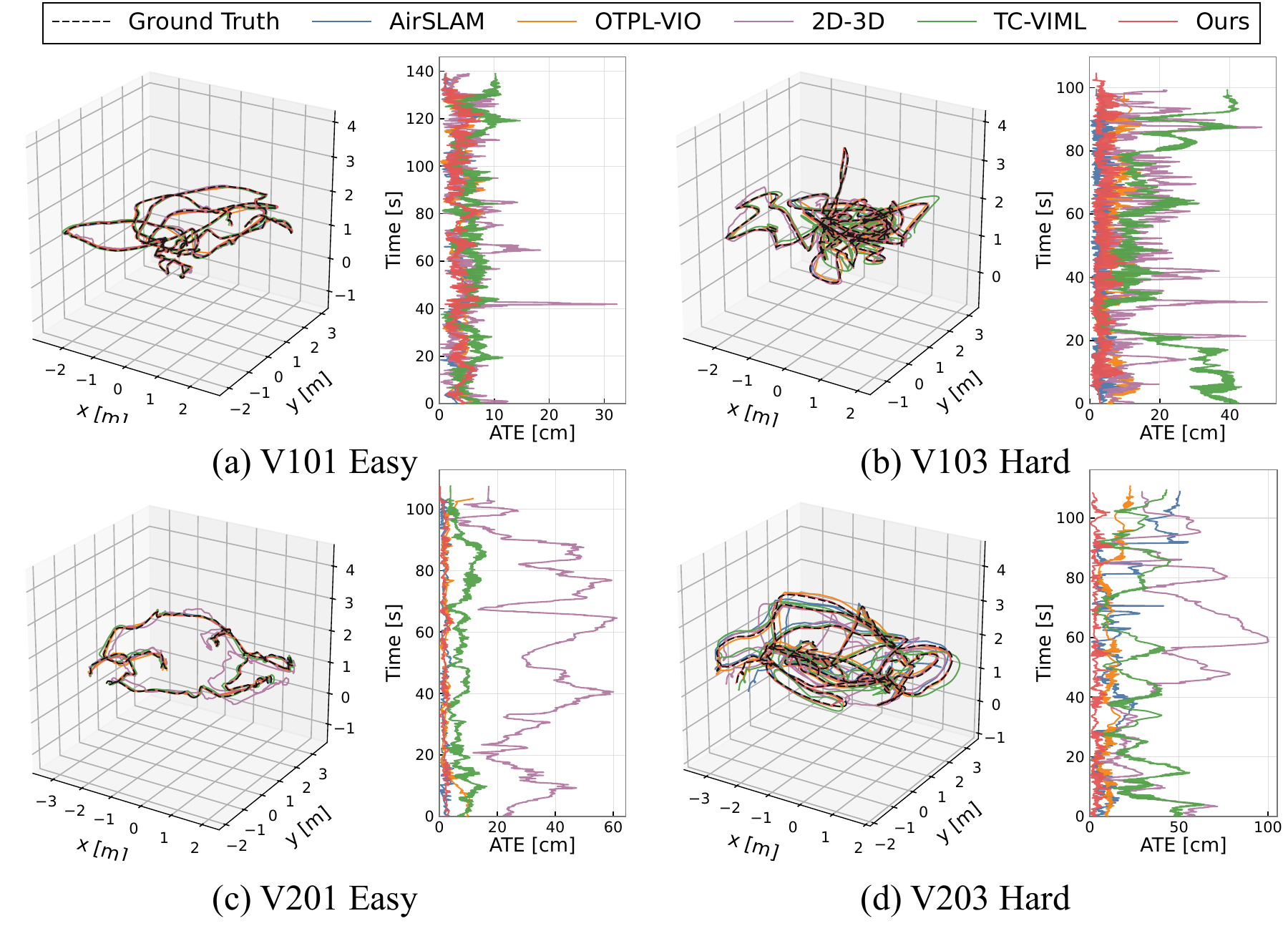}
    \caption{
    Qualitative comparison of continuous pose tracking on representative
    EuRoC sequences.
    }
    \label{fig:euroc_tracking_vis}
    \vspace{-2mm}
\end{figure}

\subsection{Evaluation on EuRoC}

\noindent\textbf{Global Localization Performance.}
Table~\ref{tab:euroc_global} reports the global localization results on EuRoC.
Following~\cite{ye20213d}, Recall denotes the percentage of queries with
translation error below $0.3$m, while mATE measures the mean translation
error of the relocalized poses.
The compared methods benefit from camera-derived reference information, such
as real reference images, visual descriptors, or additional visual traversals.
In contrast, our database is synthesized entirely from LiDAR geometry and
reflectivity, requiring direct matching between RGB queries and LiDAR-derived
virtual views without camera-derived map appearance.
Ours (Fast) searches 16 candidate groups in the first stage and expands to 32
upon failure, whereas Ours (Full) uses 40 and 200 candidates, respectively.
The former targets rapid relocalization after tracking loss, while the latter
performs a more exhaustive search for robust global initialization.
Despite the substantially more restrictive cross-modal setting, the full
configuration maintains strong recall across all four sequences, including
V103 and V203.
Meanwhile, the fast configuration retains useful localization accuracy with
substantially lower latency, as further quantified in
Table~\ref{tab:runtime}.

\noindent\textbf{Continuous Pose Tracking.}
Table~\ref{tab:euroc_tracking} reports the ATE RMSE on the EuRoC sequences,
while Fig.~\ref{fig:euroc_tracking_vis} visualizes the estimated trajectories
and frame-wise translation errors on representative sequences.
Our method achieves the best performance on most sequences and remains closely
aligned with the ground-truth trajectory throughout the evaluation.
The advantage is particularly evident on the challenging V103 and V203
sequences, where several competing methods exhibit larger trajectory
deviations and pronounced error peaks under rapid motion and limited visual
support.
Compared with recent structural-map methods such as PPL-OR and TC-VIML, the
proposed method maintains lower tracking errors by exploiting complementary
point, line, and plane constraints.
Notably, this accuracy is achieved using only a pre-built LiDAR map as the
persistent prior, without requiring camera-derived appearance in the map.

\subsection{Real-World Experiments}

\begin{figure}[t]
    \centering
    \includegraphics[width=0.8\linewidth]{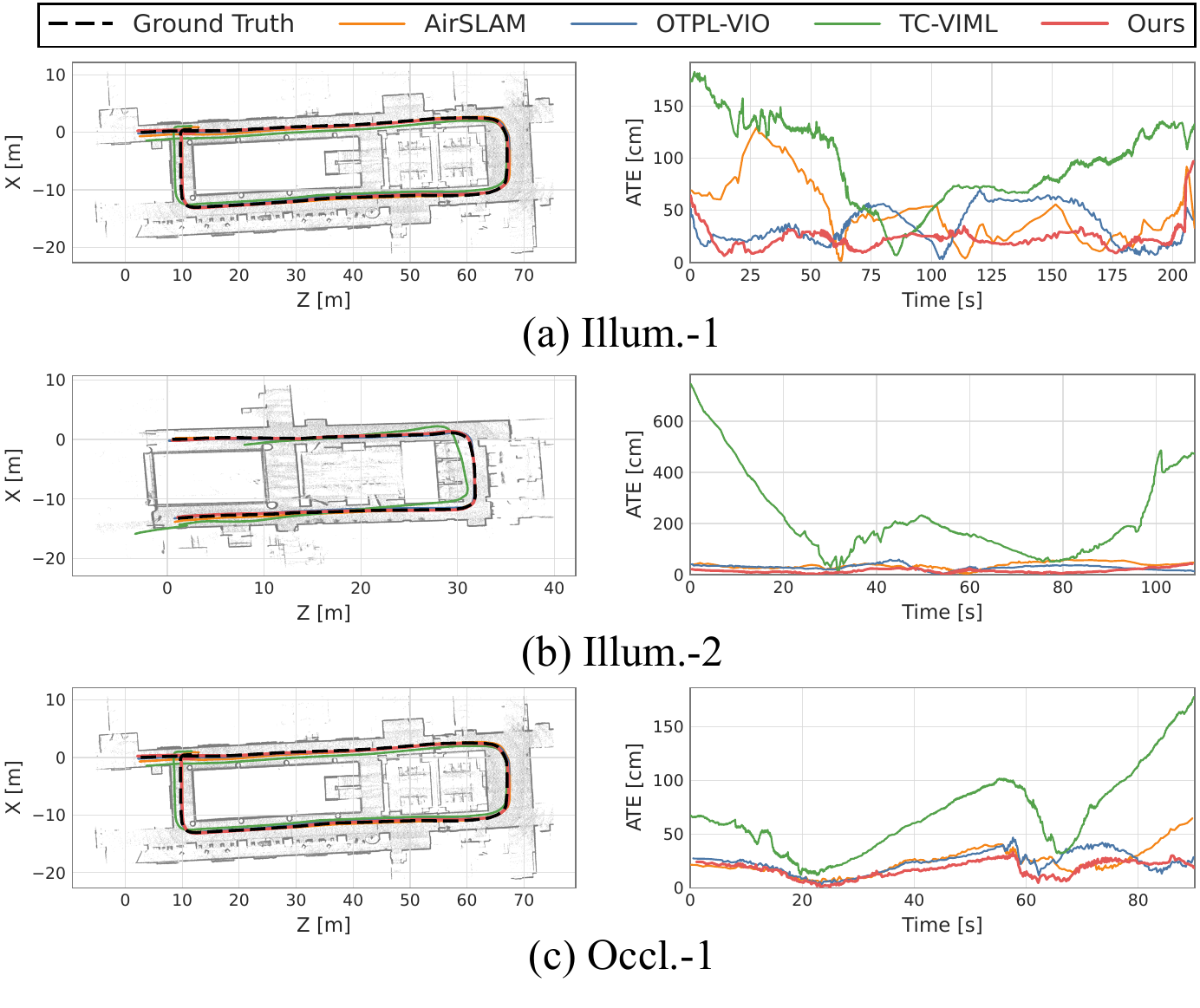}
    \caption{
    Qualitative comparison of continuous pose tracking on real-world experiments.
    }
    \label{fig:realworlds_vis}
    \vspace{-5mm}
\end{figure}

\begin{table}[t]
\centering
\caption{ATE RMSE [cm] on the real-world low-texture sequences.
\emph{Illum.} and \emph{occl.} denote illumination variation and
dynamic occlusion, respectively.}
\label{tab:real_tracking}
\setlength{\tabcolsep}{3.8pt}
\renewcommand{\arraystretch}{1.08}

\resizebox{\linewidth}{!}{%
\begin{tabular}{l|cc|ccc}
\toprule
\multirow{2}{*}{Sequence}
& \multicolumn{2}{c|}{w/o Map}
& \multicolumn{3}{c}{w/ Map} \\
\cmidrule(lr){2-3}
\cmidrule(lr){4-6}
&
AirSLAM~\cite{xu2025airslam}
& OTPL-VIO~\cite{otplvio}
& 2D--3D~\cite{yu2020monocular}
& TC-VIML~\cite{zheng2025tcviml}
& Ours \\
\midrule

Illum.-1
& 63.1 & \second{38.7} & 100.7 & 103.4 & \best{29.6} \\

Illum.-2
& 34.6 & \second{29.9} & 134.6 & 275.8 & \best{16.0} \\

Occl.-1
& 55.5 & \second{34.8} & 97.5 & 78.8 & \best{27.2} \\

\bottomrule
\end{tabular}%
}
\vspace{-3mm}
\end{table}

Table~\ref{tab:real_tracking} reports the tracking accuracy on the
self-collected real-world sequences, while
Fig.~\ref{fig:realworlds_vis} provides qualitative trajectory comparisons.
Compared with EuRoC, the self-built LiDAR priors contain less complete
structural support and larger registration and fitting noise, making reliable
line and plane constraints more difficult to obtain.
Consequently, structure-dominated map-based methods degrade when geometric
primitives are sparse or noisy, especially under strong illumination changes
and sustained pedestrian occlusions.
Our method remains more stable by combining visual point correspondences with
complementary line and plane constraints, while distribution-aware association
modeling and observability-complementary reweighting selectively reinforce
reliable structural contributions.

\begin{figure}[t]
    \centering
    \includegraphics[width=\linewidth]{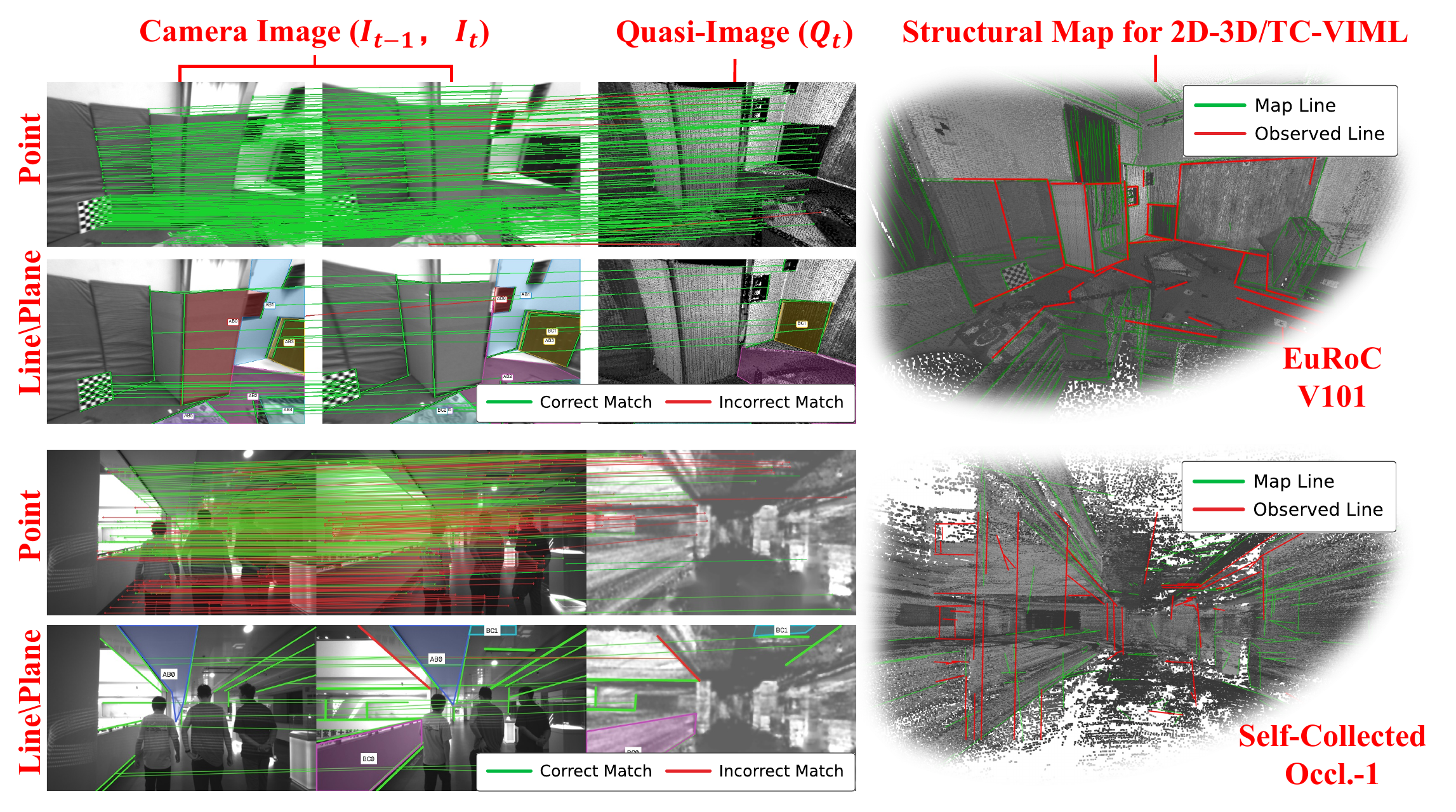}
    \caption{
    Quasi-image rendering and point/structural associations under different map
    qualities. Despite the camera--LiDAR appearance gap, consistent visual
    correspondences are established, while structural associations become sparser
    and less reliable on the self-collected maps.
    }
    \label{fig:feature_matching}
    \vspace{-5mm}
\end{figure}

\begin{table}[t]
\centering
\caption{Ablation of heterogeneous geometric primitives.}
\label{tab:ablation_plp}
\setlength{\tabcolsep}{3.0pt}
\renewcommand{\arraystretch}{1.08}
\resizebox{0.7\linewidth}{!}{%
\begin{tabular}{
l|
*{4}{>{\centering\arraybackslash}p{0.18\linewidth}}
}
\toprule
Dataset
& P
& L
& P + L
& P + L + $\Pi$ \\
\midrule
EuRoC
& 3.74 & 4.02 & 3.07 & 2.35 \\
Real-world
& 31.10 & $\times$ & 25.77 & 24.27 \\
\bottomrule
\end{tabular}
}
\end{table}

\begin{table}[t]
\centering
\caption{Ablation of observability and association modeling.}
\label{tab:ablation_confidence}
\setlength{\tabcolsep}{4.0pt}
\renewcommand{\arraystretch}{1.08}
\resizebox{0.7\linewidth}{!}{%
\begin{tabular}{
l|
*{3}{>{\centering\arraybackslash}p{0.23\linewidth}}
}
\toprule
Dataset
& Uniform
& Obs.-Aware
& Full Model \\
\midrule
EuRoC
& 5.08 & 3.78 & 2.35 \\
Real-world
& 55.21 & 35.15 & 24.27 \\
\bottomrule
\end{tabular}
}
\vspace{-5mm}
\end{table}

\subsection{Ablation and Analysis}

\noindent\textbf{Feature Association and Geometric Ablation.}
Figure~\ref{fig:feature_matching} visualizes the generated quasi-images and
the resulting point and structural associations under different map qualities.
Despite the appearance gap between camera images and LiDAR-derived
quasi-images, the shared visual front-end establishes consistent point and
line correspondences, demonstrating that the proposed representation makes the
LiDAR prior visually addressable.
Across both datasets, structural associations are inherently more selective
and sparser than point correspondences, making line and plane constraints less
widely available as standalone measurements.
On EuRoC, the accurate and complete structural prior still supports reliable
line/plane extraction and matching; accordingly,
Table~\ref{tab:ablation_plp} shows that line-only tracking remains feasible,
while combining heterogeneous PLP constraints reduces the average ATE from
3.74cm to 2.35cm.
For the self-collected sequences, mapping uncertainty, glass surfaces, local
degeneracy, and dynamic occlusions further fragment or suppress structural
primitives, substantially reducing the number and reliability of usable
line/plane associations.
In contrast, point correspondences remain comparatively more abundant and
matchable, providing a stable localization backbone.
This explains why line-only tracking fails in the real-world sequences, while
our method remains robust by exploiting point-level visual matches and using
reliable line and plane constraints as complementary geometric information.

\noindent\textbf{Effect of Observability and Association Modeling.}
We compare uniform structural weighting, observability-complementary
reweighting, and the full model with association-distribution uncertainty.
As shown in Table~\ref{tab:ablation_confidence}, observability modeling
reduces the average ATE from 5.08 to 3.78cm on EuRoC and from 55.21 to
35.15cm on the real-world sequences; incorporating association uncertainty
further reduces them to 2.35 and 24.27cm, respectively.

\subsection{Runtime Analysis}

\begin{table}[t]
\centering
\caption{Runtime analysis of the proposed framework.
Global localization is reported per query, and tracking-stage runtimes are measured on keyframes.}
\label{tab:runtime}
\setlength{\tabcolsep}{5.0pt}
\renewcommand{\arraystretch}{1.10}
\resizebox{\linewidth}{!}{%
\begin{tabular}{l|l|c|c}
\toprule
Mode & Stage & Time [ms] & Total [ms] \\
\midrule

\multirow{2}{*}{\makecell[c]{Global\\localization}}
& Retrieval (Fast/Full)
& 26.25 / 25.78
& \multirow{2}{*}{129.67 / 337.58} \\
& Geometric verification (Fast/Full)
& 103.42 / 311.79
& \\

\midrule

\multirow{4}{*}{\makecell[c]{Pose\\tracking}}
& Quasi-image generation
& 38.30
& \multirow{4}{*}{111.80} \\
& PLP extraction \& association
& 61.13
& \\
& Information modeling \& reweighting
& 0.06
& \\
& Graph optimization
& 12.33
& \\

\bottomrule
\end{tabular}
}
\vspace{-5mm}
\end{table}

Table~\ref{tab:runtime} reports the computational cost of the proposed
framework on EuRoC V103, measured on a laptop equipped with an
AMD Ryzen 9 7945HX CPU, 16GB RAM, and an NVIDIA GeForce RTX 4060 Laptop GPU.
For global localization, Ours (Full) uses the 40/200 search budget and requires
337.58ms per query, while Ours (Fast) reduces the budget to 16/32 and the
runtime to 129.67ms.
For pose tracking, keyframe refinement takes 111.80ms, with most computation
spent on quasi-image generation and PLP extraction and association.
The proposed information modeling and reweighting adds only 0.06ms.
Measured over all tracking frames, the average runtime is 75.10ms per frame,
corresponding to approximately 13.3Hz.


\section{CONCLUSION}
We presented a unified framework for global localization and continuous pose
tracking directly against a pre-built LiDAR map.
LiDAR-derived quasi-images enable shared visual association while preserving
metric 2D--3D provenance for heterogeneous PLP pose constraints.
Distribution-aware, observability-complementary optimization further
reinforces reliable structural information along weak pose directions.
Experiments on EuRoC and real-world robotic sequences demonstrate accurate
and robust localization under challenging motion, illumination changes, and
dynamic occlusions.
A remaining limitation is that the benefit of line and plane constraints
depends on sufficient structural support; under severely degraded or
structure-poor maps, localization relies predominantly on point
correspondences.

\bibliographystyle{IEEEtran}
\bibliography{main}

\newpage

\vfill

\end{document}